\documentclass[letterpaper, 10 pt, conference]{ieeeconf}  

\IEEEoverridecommandlockouts                              

\usepackage{amsmath}
\usepackage{graphicx}
\usepackage{boldline}
\usepackage{cite}
\usepackage{amsmath,amssymb,amsfonts}
\usepackage{graphicx}
\usepackage{textcomp}
\usepackage{xcolor}
\usepackage{algorithm}
\usepackage{algcompatible}
\usepackage{microtype}
\usepackage{array}
\usepackage{tikz,}
\usepackage{graphicx}
\usepackage{tabularx}
\usepackage{multirow}
\usepackage[utf8]{inputenc}
\usepackage[font=footnotesize]{subfig}
\usetikzlibrary{positioning}
\usetikzlibrary{arrows}
\IEEEoverridecommandlockouts   
\usepackage{caption}
\newcolumntype{P}[1]{>{\centering\arraybackslash}p{#1}}
\newcolumntype{M}[1]{>{\centering\arraybackslash}m{#1}}

\makeatletter
\newcommand{\removelatexerror}{\let\@latex@error\@gobble}
\makeatother

\title{\LARGE \bf
SST: A Simplified Swin Transformer-based Model for Taxi Destination Prediction based on Existing Trajectory
}

\author{Zepu Wang$^{1}$, Yifei Sun$^{2}$, Zhiyu Lei$^{1}$, Xincheng Zhu$^{1}$, Peng Sun$^{3*}$
\thanks{$^{1}$ Department of Computer and Information Science, University of Pennsylvania; {\tt\small zepu@seas.upenn.edu}; {\tt\small zlei6@seas.upenn.edu};{\tt\small kyriezxc@seas.upenn.edu}}
\thanks{$^{2}$ Stuart Weitzman School of Design, University of Pennsylvania; {\tt\small sophiasun@alumni.upenn.edu}}%
\thanks{$^{3}$ Department of Natural and Applied Sciences, Duke Kunshan University; {\tt\small peng.sun568@duke.edu}}%
\thanks{$^{*}$ Corresponding Author 
}
}

\begin{document}

\maketitle

\begin{abstract}

Accurately predicting the destination of taxi trajectories can have various benefits for intelligent location-based services. One potential method to accomplish this prediction is by converting the taxi trajectory into a two-dimensional grid and using computer vision techniques. While the Swin Transformer is an innovative computer vision architecture with demonstrated success in vision downstream tasks, it is not commonly used to solve real-world trajectory problems. In this paper, we propose a simplified Swin Transformer (SST) structure that does not use the shifted window idea in the traditional Swin Transformer, as trajectory data is consecutive in nature. Our comprehensive experiments, based on real trajectory data, demonstrate that SST can achieve higher accuracy compared to state-of-the-art methods.

\end{abstract}


\section{Introduction}
Efficient and stable transportation systems are critical to the smooth functioning of modern society, as they facilitate the movement of people and goods~\cite{wang2022novel}. Taxis are a popular mode of transportation and play an important role in the overall traffic system. However, with the rise of online ride-hailing services, traditional taxi companies are facing challenges in terms of efficient scheduling and security monitoring of their vehicles, particularly because taxi drivers cannot know their destinations in advance.

Fortunately, most taxis are equipped with mobile GPS devices that record and report their trajectories. Analyzing this trajectory data can provide insights into the destination of a taxi, which can yield several benefits such as providing location-based services and applications, alleviating traffic congestion, and optimizing taxi dispatch. At the same time, analysis of the destinations of taxi trajectories can yield several benefits such as providing alleviating traffic congestion, optimizing taxi dispatch, and location-based services and applications such as recommending sightseeing places, accurate ads based on destinations, etc.


Destination prediction based on vehicle trajectories involves using machine learning algorithms to analyze the trajectory data collected from the GPS devices installed in taxis. These algorithms can identify patterns in the data and use these patterns to predict the destination of a taxi with a high degree of accuracy. This can help taxi companies to efficiently schedule their vehicles and ensure that they are being utilized effectively.

Moreover, destination prediction can also help improve the security monitoring of taxis. By analyzing the trajectory data, it is possible to detect anomalies such as sudden deviations from a usual route or unexpected stops, which could be indicative of criminal activity~\cite{wang2022novel}.


Vehicle destination prediction is typically based on analyzing previous GPS records along with the surrounding environment, which includes factors such as the road structure and other nearby vehicles~\cite{9158529}. A variety of models have been developed to address this issue, including conventional approaches and deep learning methods.
Conventional methods such as physics-based, maneuver-based, and interaction-aware models~\cite{pepy2006reducing, gindele2010probabilistic} are limited in their ability to capture the complex spatiotemporal dependencies in the data, resulting in suboptimal prediction accuracy.
With the emergence of deep learning, researchers have explored the use of convolutional neural networks (CNNs) and recurrent neural networks (RNNs) for trajectory prediction~\cite{lv2018t, deo2018convolutional}. These methods leverage the power of deep learning to capture non-linear relationships and long-range dependencies in the data, resulting in significant improvements in prediction accuracy.
More recently, graph-based techniques such as graph convolutional networks (GCNs)~\cite{kipf2016semi} have been incorporated to model the spatial structure and interactions between taxis within the road network. GCNs can effectively model the underlying structure of road networks and capture the interactions between different taxis, leading to improved prediction accuracy.

In order to fully capture the spatial information of a trajectory, researchers often convert it into a two-dimensional map since it is highly related to the structure of road networks. This allows for the utilization of more advanced computer vision techniques to solve prediction problems. In recent years, with the breakthroughs in computer vision using vision transformers, many scholars have been inspired to use them for trajectory or destination prediction and have achieved good results~\cite{zhao2021spatial,liu2021multimodal}. Furthermore, Swin-transformer, a variant of vision transformers, has become a general-purpose backbone for computer vision tasks~\cite{liu2021swin}. However, to the best of our knowledge, the Swin architecture has not been widely used in trajectory analysis or destination prediction before.

The main contributions of this paper are as follows:
\begin{itemize}

\item[$\bullet$] Firstly, an SST is proposed that is better suited to the destination prediction problem. This model is shown to be competitive for spatiotemporal prediction of taxi destinations, providing a new perspective for researchers seeking to apply state-of-the-art computer vision techniques to destination prediction problems.

\item[$\bullet$] Secondly, the study compares three grid-based modeling approaches for destination prediction and evaluates their effectiveness in fitting traditional trajectory data into a trajectory grid. The results of this comparison can provide insights into the an effective way to convert traditional trajectory data into trajectory grids for further analysis.

\end{itemize}

The remainder of this paper is structured as follows. Section~\ref{Literature Review} provides a comprehensive review and summary of previous studies related to trajectory prediction. Section~\ref{Problem Statement} describes the problem statement. It discusses how travel trajectory data is collected from taxis once they start carrying passengers. The proposed methodology, including data processing and model structure, is described in Section~\ref{Proposed Method}. Section~\ref{EXPERIMENT} presents experimental results that compare several models and their performance against our proposed approach. The results demonstrate the effectiveness of our SST in predicting taxi destinations based on trajectory data. In Section~\ref{Conclusion}, we summarize the contributions of this study and highlight its potential impact on the transportation system and society as a whole. 
  
\section{Literature Review}
\label{Literature Review}
 
Trajectory analysis is widely studied in the literature using traditional and machine learning (deep learning) approaches. Early studies in trajectory prediction employed physics-based models such as dynamic models \cite{lin2000vehicle, pepy2006reducing} and kinematic models \cite{ammoun2009real, kaempchen2004imm}, which predict future vehicle motion based on vehicle attributes, control inputs, and external factors such as the vehicle's position, heading, and speed. While physics-based models are widely used in trajectory prediction and collision risk estimation, their ability to predict trajectories over a long time is limited by their reliance on low-level motion properties.

In contrast, some researchers propose maneuver-based models that consider prior knowledge, making them more reliable than physics-based models. These models are based on prototype trajectories \cite{vasquez2004motion} or maneuver intention estimation \cite{klingelschmitt2014combining, berndt2008continuous}. However, they do not consider external objects such as surrounding vehicles, which can cause misjudgments. To address this limitation, interaction-aware models were developed that treat vehicles as maneuvering entities that can be affected by other vehicles in a scene. These models are based on either prototype trajectories \cite{kafer2010recognition} or Dynamic Bayesian Networks \cite{agamennoni2012estimation}, and show better results than traditional maneuver-based models. Nonetheless, these models suffer from expensive computation problems, as they need to compute all possible vehicle trajectories.

In recent years, with the advancement of deep learning, learning-based techniques are increasingly applied to solve vehicle trajectory prediction problems. As trajectories possess sequential attributes, the problem can be addressed as a time-series prediction task. Therefore, many scholars utilize typical recurrent neural networks (RNNs) \cite{rumelhart1986learning}, long short-term memory (LSTM) neural networks\cite{hochreiter1997long}, and gated recurrent unit (GRU) networks\cite{chung2014empirical} as their basic structures to design the model. For instance, Kim et al.\cite{kim2017probabilistic} propose an LSTM-based framework to learn various behaviors of vehicles from massive trajectory records. Deo et al.\cite{deo2018multi} propose an LSTM model for trajectory prediction under the scene of the freeway, which not only includes track histories but also takes into account surrounding vehicles and road structures as input. Lee et al. use the RNN Encoder-decoder framework to build the DESIRE model\cite{lee2017desire}, which accurately predicts the future locations of objects across various scenes.

Most existing methods for trajectory prediction focus on modeling trajectories as a one-dimensional time series, which may not fully capture the complex nonlinear spatial-temporal correlations inherent in trajectory data. This limitation becomes particularly evident when predicting trajectories that are highly related to road structures, such as those involving corners or winding paths. To overcome this limitation, some recent works propose to transform trajectory data into a two-dimensional matrix format, where each pixel corresponds to a specific location and encodes information about the presence and movement of vehicles in that location over time. Therefore, more computer vision architectures such as convolutional neural networks (CNNs) can be used to extract more spatial information and build relationships with surrounding objects. For instance, Lv et al. propose a CNN-based model \cite{lv2018t} that takes vehicle trajectory prediction as an image prediction task and combines multi-scaled trajectory patterns. The model shows high accuracy in trajectory prediction tasks. Similarly, Guo et al. combine CNN with LSTM \cite{xie2020motion} to predict the trajectory of surrounding vehicles by merging the spatial expansion properties of CNN and the temporal expansion capabilities of LSTM. The model shows better performance than using time-series models such as LSTM or GRU alone.


It is worth noting that since 2017, attention-based/transformer-related models demonstrate impressive performance in various application scenarios. In the field of computer vision, the Vision Transformer (ViT) proposed by Dosovitskiy et al. \cite{9716741} has shown remarkable results in many computer vision tasks. In our research, we employ a more specific type of transformer, namely the Swin Transformer \cite{Liu_2021_ICCV}. The Swin Transformer generates hierarchical feature maps by merging image patches in deeper layers, and its linear computation complexity to input image size is due to the computation of self-attention only within each local window. As a result, it can function as a general-purpose backbone for both image classification and dense recognition tasks. A comprehensive description of our modified Swin structure is provided in Section~\ref{Proposed Method}.

\section{Problem Statement}
\label{Problem Statement}
When taxis start carrying passengers, it is able to begin collecting the travel trajectory of occupied taxis. Specifically, given a taxi $X_i$, its $j$-th trajectory $Y_{ij}$ is recorded as a sequence of GPS locations in a fixed period: $Y_{ij} = \langle l_{ij1}, l_{ij2}, l_{ij3}, \ldots, l_{ijN_{ij}}\rangle$. Each $l_{ijk}$ in the sequence represents a GPS location of longitude and latitude pair $(A_{ijk}, B_{ijk})$, collected instantly, where $1 \leq k \leq N_{ij}$
. Here, $N_{ij}$ is the total length of the taxi's current trip path $Y_{ij}$. It is worth noting that the total length of different trajectories can vary.

To clarify, $l_{ij1}$ is the start of the trip where the taxi takes on the passenger(s). The destination of the taxi is represented by the last location in the sequence, $l_{ij, N_{ij}}$. We define the destination of any trajectory $Y_{ij}$ as $\zeta_{Y_{ij}}$. Our prediction problem can be defined as predicting the final destination $\zeta_{Y_{ij}}$ of a taxi $X_i$ in a trip $Y_{ij}$, given its historical trajectory set.

\section{Proposed Method}
\label{Proposed Method}
\subsection{Grid-based modeling of trajectory}
\begin{figure}[htp]
    \centering
    \includegraphics[width=0.45\textwidth]{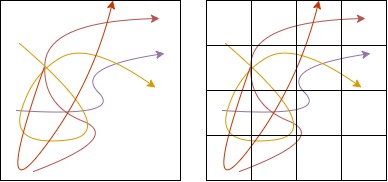}
    \caption{An illustration of two-dimensional grid representation}
    \label{two d image}
\end{figure}

To extract spatial patterns from the taxi trajectories, a two-dimensional grid representation is adopted. Then, divide the map into an $M\times M$ grid, where $M$ is a predefined constant resolution of the map. Each GPS location $l_{ijk}$ is mapped onto a corresponding grid cell $G_{pq}$ based on its latitude and longitude, where $m$ and $n$ represent the row and column indices of the grid cell, respectively. This mapping relationship is denoted as $l_{ijk} \rightarrow G_{mn}$. By applying binary, linear, or quadratic transformation methods, the pixel values $I_{ij}(m,n)$ (where $1 \le m,n \le M$) of the resulting $M\times M$ matrix $I_{ij}$ can be defined. These matrices serve as the two-dimensional grid representation of the taxi trajectories. Here, we present three methods to make the taxi trajectories. In Figure~\ref{two d image}, different arrows represent 4 different trajectories and it is displayed in a $4\times 4$ matrix.

\subsubsection{Binary Method}
\begin{equation}
I_{ij}^{\mathrm{bin}}(m,n)=\left\{
    \begin{aligned}
    1, & \mathrm{\ if\,}\exists l_{ijk}, (l_{ijk} \in Y_{ij} \wedge l_{ijk} \rightarrow G_{mn})\\
    0, & \mathrm{\ else}
        \label{binary transform equation}
    \end{aligned}
    \right.
\end{equation}
\par
In the binary method, a value of 1 is assigned to a grid cell if the taxi passes through the area where the grid cell represents at any given time, and a value of 0 is assigned otherwise. However, this method only captures the track of the vehicle and does not account for temporal information.
\subsubsection{Linear Method}
\begin{equation}
I_{ij}^{\mathrm{lin}}(m,n)=\left\{
    \begin{aligned}
    k/N_{ij}, & \mathrm{\ if\,}\exists l_{ijk},(l_{ijk} \in Y_{ij} \wedge l_{ijk} \rightarrow G_{mn})\\
    0, & \mathrm{\ else}
        \label{linear transform equation}
    \end{aligned}
    \right.
\end{equation}
\par
In the linear method, the temporal dimension of the trajectory is taken into account by equally dividing the time into $N_{ij}-1$ parts, where $N_{ij}$ is the total length of the trajectory. As time passes during the taxi trip, the pixel value representing the location of the taxi gradually increases from 0 to 1, providing an intuitive way of encoding the temporal information in the trajectory grid.

\subsubsection{Quadratic Method}
\begin{equation}
I_{ij}^{\mathrm{qua}}(m,n)=\left\{
    \begin{aligned}
    (k/N_{ij})^2, & \mathrm{\ if\,}\exists l_{ijk}, (l_{ijk} \in Y_{ij} \wedge l_{ijk }\rightarrow G_{mn})\\
    0, & \mathrm{\ else}
        \label{quadratic transform equation}
    \end{aligned}
    \right.
\end{equation}
\par

In the quadratic method, the pixel values of the linear method are quadratically transformed at each time step. This approach allows for a more nuanced representation of the temporal information in the trajectory grids. Specifically, if the final destination is more strongly correlated with the later portions of the trajectory and less with the initial locations, the quadratic method assigns relatively higher values to the later portion of the trajectory. As a result, the quadratic processed matrix exhibits more prominent features associated with the later portion of the trajectory compared to those processed using the linear method.

\subsection{Simplified Swin Transformer (SST)}

Inspired by the successful application of the Swin transformer~\cite{liu2021swin} in many computer vision tasks, we design an SST for the destination prediction task.

\begin{figure*}
    \centering
    \includegraphics[scale=0.5]{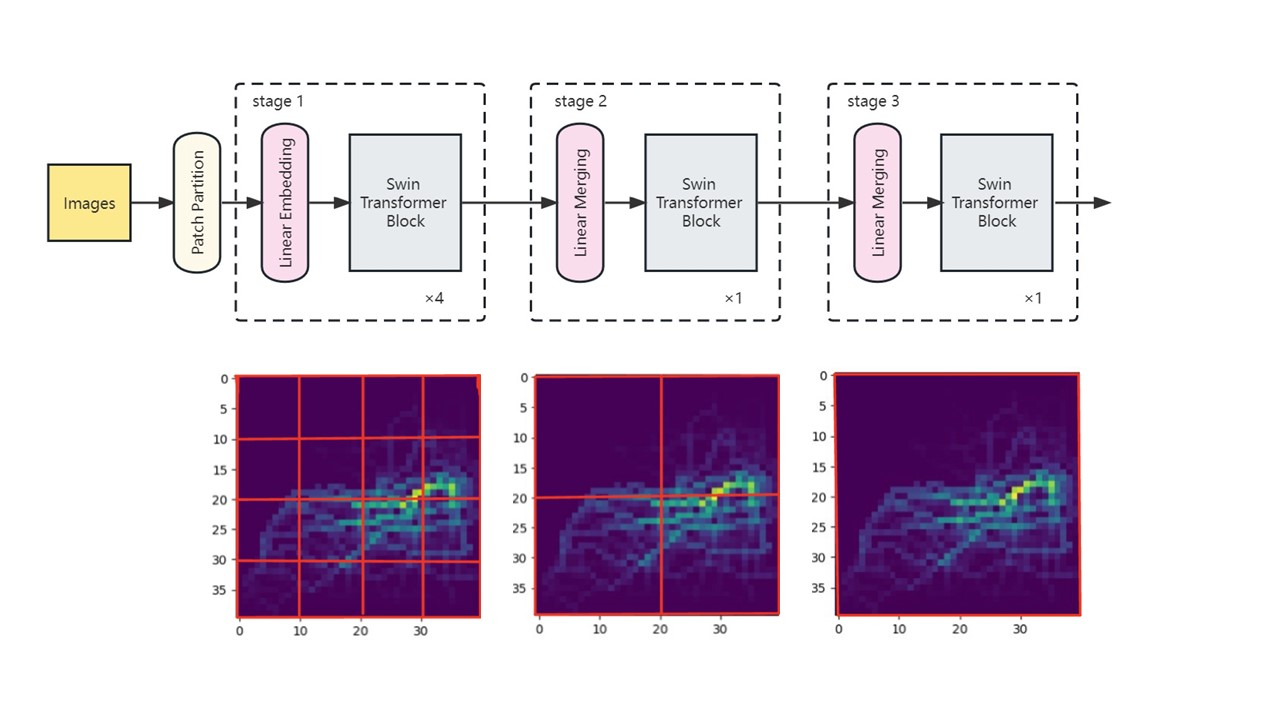}
    \caption{The architecture of the SST}
    \label{SwinT}
\end{figure*}

\subsubsection{General Structure}

An overview of the Swin Transformer architecture, which has been adopted in this paper for trajectory prediction, is presented in Figure~\ref{SwinT}. The architecture first splits the input trajectory grid into non-overlapping patches using a patch-splitting module. However, unlike traditional image inputs, the trajectory matrix input only has one channel. In this implementation, we use a patch size of 10 x 10 in the patch partition stage.

After processing by the first stage, in order to produce a hierarchical representation, the network merges small patches into bigger ones as it goes deeper. As illustrated in the figure, the first patch merging layer concatenates the features of each group of 2 × 2 neighboring patches. Since the trajectory input size in our experiment is 40, after the second merging, the network calculates the global attention to the big matrix. In Figure~\ref{SwinT}, we illustrate how the architecture calculates attention scores in different stages. The picture is the mean of all the trajectories that we use in Section~\ref{EXPERIMENT}. In this way, we can capture the global relationships between different pixels without having to calculate the global attention multiple times, which could waste computational resources.

The Swin Transformer architecture has several key advantages to our trajectory prediction task. It allows for efficient computation and scalability to handle larger grid matrices by using a hierarchical patch-based approach, which can be applied to the trajectory grid data. Additionally, the use of attention mechanisms in the architecture allows for the model to better capture long-range dependencies between pixels, which is important for accurately predicting the final destination of a taxi trajectory.



\subsubsection{SST block}

\begin{figure}
    \centering
    \includegraphics[scale=0.5]{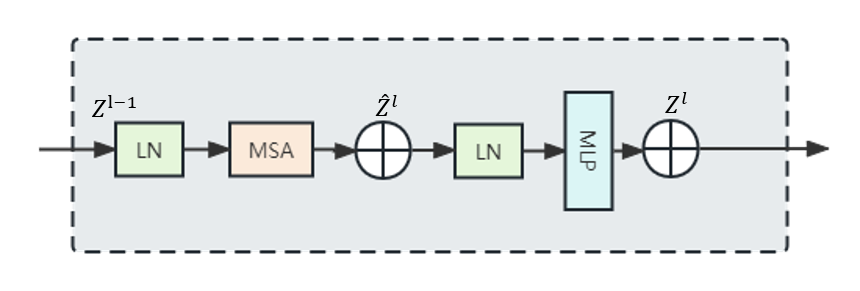}
    \caption{The architecture of the SST block}
    \label{block}
\end{figure}

Figure~\ref{block} provides an illustration of the SST block. The input is first processed by a Layernorm (LN) layer, followed by a conventional multi-head self-attention (MSA) module. MSA is an extension of self-attention, which is a technique used to calculate the importance of different parts of the input when predicting an output\cite{vaswani2017attention}. In MSA, we run $k$ self-attention operations, in parallel, and combine their outputs together. The output then undergoes processing by an LN layer and MLP layer, before being subject to another residual connection.

In contrast to the traditional Swin Transformer, which proposes computing self-attention within local windows and applying a shifted window partitioning approach to enhance connections between non-overlapping windows, this paper argues against using shifted window-based MSA. This decision is based on the nature of trajectory input, which represents the continuous change in a vehicle's state within the real world, with the velocity of vehicles limited to a specific range. Therefore, for a specific pixel, it is more important to calculate the attention between it and some close pixels than between some far pixels.

Using a trajectory matrix as an input implies that a specific pixel representing the vehicle's location at a given time should be more related to its few previous states and next several locations than pixels further away. Hence, applying the shifted window technique in this work would result in the network finding relationships between two patches located far from each other, even though such a relationship should not exist. Therefore, our SST block adopts a more straightforward approach to self-attention, avoiding the shifted window technique and focusing on capturing local and global dependencies in a way that is better suited to the characteristics of trajectory matrix input.

\section{EXPERIMENT}
\label{EXPERIMENT}
\subsection{Data Preparation}
In our experiment, we evaluate the performance of the SST model along with other baseline models on a real trajectory dataset from the ECML-PKDD competition \cite{kaggle}. The dataset comprises 1.7 million complete trajectories collected from 442 taxis that operated in the city of Porto for a year, from 2013-07-01 to 2014-06-30. Each trajectory consists of a list of longitude and latitude pairs that represent the recorded positions of a taxi during the trip. To reduce the dataset's size, we randomly sample 100,000 trajectories. As the city of Porto is vast, we only retain trajectories within a certain longitude and latitude range ($[-8.7,-8.6]$ and $[41.1,41.2]$ respectively) since most of the trajectories fall within this range.

Next, we apply MIN-MAX normalization to map all the longitude and latitude values to the range of $[0,1]$. We then convert these values to their corresponding pixel coordinates on an $M \times M$ grid, where $M$ is chosen to be 40 after several trials and errors. Finally, we use the last coordinate in the list as the target for the prediction task and the remaining coordinates as input to construct the trajectory input.

After these preprocessing steps, we randomly split the dataset into three sets: 60\% for training, 20\% for validation, and 20\% for testing. 



\subsection{Evaluation Metrics}

In this experiment, the problem is approached as a regression task, in which the final destination is predicted based on the trajectory matrix data. The mapping relationship also processes the predicted coordinates of the destination. To evaluate the performance of the different models, the mean square error ($MSE$) is used as the primary evaluation metric.

The $MSE$ is defined as the average of the squared differences between the predicted values and the ground truth values, and is expressed as follows:

\begin{equation}
MSE = \frac{1}{n}\sum_{i=1}^n(y_i-\hat{y_i})^2,
\end{equation}

where $y_i$ and $\hat{y_i}$ denote the ground truth and predicted values, respectively, and $n$ is the total number of samples.

In this experiment, the mean absolute error ($MAE$), which is another commonly used evaluation metric, is not utilized. This decision is based on the observation that the dataset is preprocessed and cleaned, and therefore, there are no significant outliers in the data. As such, the MSE is preferred over the MAE, as it tends to penalize larger differences between the predicted and actual values more severely, which is desirable in this context.

\subsection{Result Evaluation}
We adopt three baseline models in this experiment:
\begin{itemize}

\item[$\bullet$] Multilayer perceptron (MLP): A simple multi-layer perceptron with four hidden layers of 150 neurons followed by a dropout layer.

\item[$\bullet$] Convolutional neural networks (CNNs): A simple convolutional neural network with a $7\times7$ convolution kernel with 128 channels followed by two fully connected layers.

\item[$\bullet$] Long short-term memory neural networks (LSTMs): Process the sequential input directly without the trajectory input. The list of pixel coordinates is first converted to a $200\times3$ tensor as the model input. The 200 rows represent a sequence with fixed length 200: for a list with length $L$ greater than 200, only take the last 200 coordinates; for $L$ less than 200, the last $L$ rows of the tensor are filled with coordinates while the rest are padded with zeros. The three columns indicate three input features: the first two are the pixel coordinates from the list while the third is the constant one for non-zero-padded rows. The LSTM model has one recurrent layer and hidden states with four features. The final output layer is used to map the last hidden state in the sequence to the target coordinate.

\end{itemize}

\begin{table}[h]
\caption{Mean Square Errors on Test Set}
\label{result}
\begin{center}
\begin{tabular}{V{5}c||c|c|cV{5}}
\hlineB{5}
Model\ \textbackslash\ Method & Binary & Linear & Quadratic\\
\hlineB{3}
MLP & 6.8748 & 3.2009 & 2.6359\\
\hline
CNN & 5.1156 & 1.7879 & 1.5543\\
\hline
SST & 5.2952 & 1.7722 & \textbf{1.4865}\\
\hline
LSTM & \multicolumn{3}{|cV{4}}{1.6562} \\
\hlineB{5}
\end{tabular}
\end{center}
\end{table}

Table~\ref{result} presents the experimental results obtained by employing three types of grid-based modeling of trajectory definition. Our findings suggest that the SST transformer model with the quadratic trajectory preprocessing method achieves the lowest $MSE$ error of 1.4865.

The quadratic method outperforms the binary and linear methods in all MLP, CNN, and SST models. This result aligns with our expectations since the quadratic method captures more prominent features associated with the later portion of the trajectory, resulting in better performance of prediction methods. In contrast, the binary method loses sequential information during the trajectory grid conversion, thereby preventing the CNN and SST models from capturing vehicle direction in the trajectory. While the performance of linear and quadratic methods is similar, the slightly better performance of the quadratic method indicates that it can more accurately predict the final destination. Hence, we use experiments to prove that the quadratic processed grids exhibit more prominent features associated with the later portion of the trajectory compared to those processed using the linear method.

\begin{figure}
    \centering
    \includegraphics[scale=0.5]{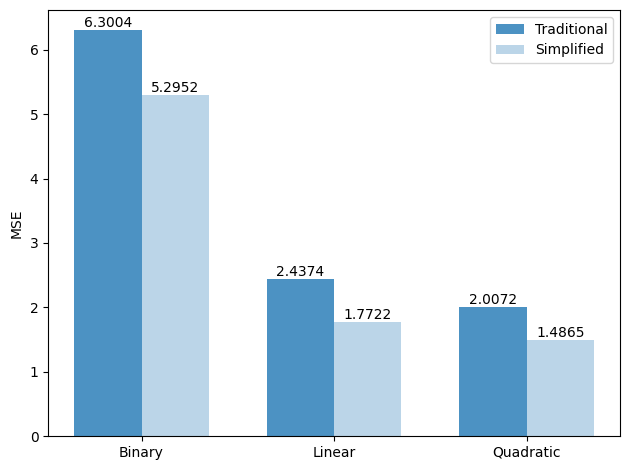}
    \caption{Comparison between traditional and SST}
    \label{comparison}
\end{figure}

Additionally, we also compare the performance of the traditional Swin transformer and our modified simple version in Figure~\ref{comparison}. For all binary, linear, and quadratic methods, SST has a higher accuracy. Therefore, to some extent, we demonstrate that the shifted window technique is not appropriate for this destination prediction problem.

Interestingly, our experiment also find that the LSTM model shows relatively good performance compared to most other experiments. This finding suggests that LSTM is an ideal trajectory analysis technique, despite the fact that it requires the length of the trajectory to be the same in order to train the LSTM model. However, since real-world trajectories often have different lengths, embedding the various trajectory sequences into a common embedding space may be a promising research direction for building more complex LSTM-related models.


\section{Conclusion}
\label{Conclusion}

This paper proposes a novel approach to predict the destinations of taxi trajectories, which involves three different trajectory grid formation methods and the use of a simplified Swin transformer (SST) model. Our results show that the quadratic method is the most effective technique for this task, while also demonstrating that the SST outperforms the traditional version, indicating that the shifting window technique is not necessary for trajectory analysis.

In the future, further experiments with additional datasets could be conducted to evaluate the performance of the SST more comprehensively. Moreover, as highlighted in Lv et al.~\cite{lv2018t}, different portions of a trajectory may have varying contributions to the final prediction. Thus, exploring the use of trajectory processing methods that can leverage these differences could be a fruitful avenue for future research.

\section*{Acknowledgment}
This work is supported by the National Natural Science Foundation of China (Grant No. 62250410368).

\bibliographystyle{IEEEtran}
\bibliography{citation}
\end{document}